# Combining model tracing and constraint-based modeling for multistep strategy diagnoses

Gerben van der Hoek[1][0009-0004-0932-3065], Johan Jeuring[1][0000-0001-5645-7681],

and Rogier Bos[1][0000-0003-2017-9792]

[1] Utrecht University PO Box 85.170, 3508 AD Utrecht, the Netherlands
g.vanderhoek@uu.nl

**Abstract.** Model tracing and constraint-based modeling are two approaches to diagnose student input in stepwise tasks. Model tracing supports identifying consecutive problem-solving steps taken by a student, whereas constraint-based modeling supports student input diagnosis even when several steps are combined into one step. We propose an approach that merges both paradigms. By defining constraints as properties that a student input has in common with a step of a strategy, it is possible to provide a diagnosis when a student deviates from a strategy even when the student combines several steps. In this study we explore the design of a system for multistep strategy diagnoses, and evaluate these diagnoses. As a proof of concept, we generate diagnoses for an existing dataset containing steps students take when solving quadratic equations ($n = 2136$). To compare with human diagnoses, two teachers coded a random sample of deviations ($n = 70$) and applications of the strategy ($n = 70$). Results show that that the system diagnosis aligned with the teacher coding in all of the 140 student steps.

## 1 Introduction

Tutoring systems employ various approaches to automatically assess student input and provide feedback. For instance, large language models (LLMs) can provide feedback without human authoring (Baral et al., 2024). The STACK environment for mathematics (Sangwin, 2015) compares properties of an input to desired properties. Constraint-based modeling (CBM) (Mitrovic, 2012) checks whether or not input satisfies one or more constraints. Model tracing (MT) (Anderson et al., 1995) uses production rules to model student steps in a domain.

The part of an assessment system that provides a domain-specific diagnosis is called a domain reasoner. The IDEAS framework for developing domain reasoners is state-of-the-art and has been used in many tutoring systems (Gerdes et al., 2012, 2017; Heeren & Jeuring, 2020; Tacoma et al., 2020; Van der Hoek et al., 2024). One goal of the IDEAS diagnosis service is to provide a diagnosis if a student deviates from a model strategy. It does so by connecting consecutive student inputs using a single production rule. If a student combines two or more steps into one step but still follows the correct strategy, the service diagnoses this as: "Correct rewrite step, but unknown".

Bokhove & Drijvers (2012) studied strategy feedback in an algebra tutor using the IDEAS framework. They report situations in which a student does not receive adequate feedback because of combining steps. The logs from the experiments in their study show that in at least 80% of the "Correct rewrite step, but unknown" diagnoses a student did follow a correct strategy but combined several steps. In this paper, we elaborate on this by adapting the diagnosis service to also allow a student to combine two or more steps. Therefore, we investigate the following research question: How can we design a service that provides a strategy diagnosis when a student combines several steps? We introduce an approach to provide a strategy diagnosis when a student combines two or more steps.

The key idea of the approach is to identify variations on a model solution through constraints. If a student deviates from a strategy, a violated constraint provides information about the way the student deviated. Our approach combines elements from MT and CBM. To validate our approach, we apply it to logs also used in the study of Bokhove & Drijvers (2012) and let two teachers assess the quality of a random sample of the diagnoses. In the following section we elaborate on the MT and CBM paradigms and explain how expert strategies can be embedded in an MT domain reasoner. We then zoom in on the IDEAS framework for implementing domain reasoners.

## 2    Theoretical framework

The IDEAS framework is an MT variant (Heeren et al., 2010; Heeren & Jeuring, 2017) that encompasses a domain-specific language (DSL) for stepwise exercises, and normal forms that describe the structure and representation of objects in a knowledge domain (e.g., quadratic equations) (Heeren & Jeuring, 2009). Expert knowledge can be embedded into a domain reasoner in various ways, for instance in the form of pairs of constraints by CBM. The first constraint, the so-called relevance condition, determines whether a situation is such that the second constraint, the satisfaction condition, could apply. For example, if a student's task is to solve $a\,x = b$ for $x$, we can define a relevance condition as "the student input is of the form $x = k$", and a satisfaction condition as "$k$ should be of the form $b/a$". If the relevance condition holds but the satisfaction condition does not, an error is flagged.

MT embeds expert knowledge through production rules. The rules transform an object and guide a student in performing the same transformation. For example, if a student is solving $a\,x = b$, one rule would be "$a\,x = b \rightarrow x = b/a$". If a student input does not match any object expected by application of a rule, an error, or deviation from the model solution, can be flagged. Table 1 shows some key differences between CBM and MT (Mitrovic et al., 2003). The rows on "problem solving strategy" and "problem solving hints" illustrate that CBM can diagnose any input, even if a student combines steps, and MT can diagnose the strategy a student follows. Various studies report on successful implementations of a combination of CBM and MT utilizing advantages of both paradigms (Roll et al., 2010; Tacoma et al., 2020).

Table 1. Comparison of model tracing and constraint-based modeling (Mitrovic et al., 2003)

| Property | Model tracing | Constraint based modeling |
| --- | --- | --- |
| Problem solving strategy | Implemented ones | Any strategy |
| Problem solving hints | Yes | Only on missing elements but not on strategy |
| Cognitive fidelity | Tends to be higher | Tends to be lower |

The level of congruence between the way a person reasons about a task, and the way knowledge is implemented in a domain reasoner is called cognitive fidelity (Mitrovic et al., 2003; Ohlsson & Mitrovic, 2007). Table 1 shows that cognitive fidelity is generally higher for MT than for CBM.

In the IDEAS framework for domain reasoners, MT is implemented by building an expert strategy from production rules and combinators defined in the DSL (Heeren & Jeuring, 2017). This is similar to building a regular expression where the symbols are the production rules using combinators such as concatenation, choice, and Kleene star. We call such an expression in the DSL a DSL-expression or strategy. Analogous to how a regular expression can generate sentences in a language, a DSL-expression can generate sequences of production rules. These sequences can be applied to a starting expression to derive model solutions for the task given the strategy and starting expression. We say that the model solutions are generated by applying the strategy to the task.

Variations on a model solution can be implemented by using normal forms of expressions. A normal form is a function $f$ that provides a canonical representation of an object. An example of such a normal form is given by simplifying fractions, here $f(n/d) = n'/d'$ where $n' = n/\gcd(n,d)$ and $d' = d/\gcd(n,d)$. Through normal forms, different representations of an object can be compared (Sangwin, 2013). Heeren & Jeuring (2009) use normal forms to describe how a single production rule can be applied on several variations of expressions. In our approach normal forms play an important role.

## 3 Identifying variations

We are interested in tasks for which a model solution exists: a solution following a strategy that the teachers or expert deems optimal in some sense. However, a student may construct a correct solution using a variant of this ideal model solution. We distinguish three types of variations and provide examples. The first type of variation occurs when an expression can be represented in various ways. For example, any order of terms is considered adequate for a quadratic equation. The second type occurs when there are choices in the order of comparable strategy steps. An example is the order of intermediate steps in which similar terms can be taken together in an equation. The third type occurs when an extra step is performed to tidy up an expression before continuing.

These variations often lead to many different solution paths and deciding whether or not a student follows an expert solution method when a student combines steps can be costly due to combinatorial explosion (Ohlsson & Mitrovic, 2007). Our solution to this problem is to only use production rules for steps in the model solution and to identify a variation on a step in a student solution through normal forms and relations. This reduces the number of production rules in a DSL expression, making multi-step diagnosis manageable. We call this approach Property Tracing (PT).

We illustrate PT with the following example. Consider the equation $(-x + 1)^2 = 9$. A way of solving this equation is to fully expand the square, derive the equation to zero and use the quadratic formula. However, the preferred strategy is to take the square root on both sides of the equation yielding the following model solution: $S = \{[(-x + 1)^2 = 9], [-x + 1 = 3, -x + 1 = -3], [-x = 2, -x = -4], [x = -2, x = 4]\}$. Suppose we wish to teach a student to use this strategy instead of the strategy of expanding the square. To diagnose student input in this example, we use three relations in the following order: (1) Expected normal form: this relation checks whether two equations have the same normal form that derives the equation to zero and simplifies as much as possible while retaining the factorization (i.e., the bracketed structure); (2) Expected number of terms: this relation checks whether two expressions have the same number of non-zero terms; (3) Expected zero derivation: checks whether two equations are both derived to zero or are both not derived to zero.

Suppose a student enters: $(-x + 1)^2 - 9 = 0$. We then use the relations above to check whether there is an element in $S$ that satisfies all relations with this input. In this case, the student input has the same normal form as: $(-x + 1)^2 = 9$, this also has the same number of non-zero terms as the student input, but it is not derived to zero in the same way. In this case some kind of feedback that mentions that the student unexpectedly derived to zero can be presented to the student.

Suppose a student enters: $x^2 - 2x - 8 = 0$. Here relation (1) is violated and feedback mentioning an unexpected application of the distribution law can be given. The order of checking the relations is important. Suppose we first check relation (3) in the above example. A violation of relation (3) results in feedback about zero derivation. Subsequently the student might write $x^2 - 2x = 8$, which is not helpful. Therefore, in this case relation (3) only becomes relevant once relation (1) and (2) are satisfied.

Suppose a student enters: $[1 - x = 3, 1 - x = -3]$. Although this is not part of the model solution, it does satisfy relation 1,2 and 3 with $[-x + 1 = 3, -x + 1 = -3]$, and it is recognized as a variant of the model solution. In this case positive feedback can be presented to the student. Moreover, a strategy hint can be provided based on the rules in the DSL-expression that apply to $[-x + 1 = 3, -x + 1 = -3]$.

## 4     Evaluation of the PT-design

In this section we elaborate on the participants, the instrument, data collection, data analysis, and the results of an evaluation of the PT-design for quadratic equations.

*Participants*: Our data holds data from the research by Bokhove & Drijvers (2012). Their intervention took place in 2010 in fifteen 12th-grade classes from nine Dutch secondary schools ($n = 324$), with students of age 17 or 18. The schools were spread across the country and varied in size and pedagogical and religious backgrounds. The participating classes were of pre-university level. Of the participants 43% were female and 57% male.

*Instrument*: Students were presented with the task of solving quadratic equations. The tasks were embedded in the Digital Mathematics Environment (Bokhove, 2017) equipped with an IDEAS domain reasoner. In this learning environment, a student could enter a solution in a stepwise fashion and receive feedback.

*Data collection*: We used the data on solving quadratic equations of 18402 diagnosis requests of mathematically correct steps. We removed the steps containing a final answer, because a correct final answer is always part of the expert strategy, leaving 6455 requests. Of these requests, IDEAS returned a diagnosis 'unknown' for 2048 requests (31,7%). Our dataset consists of these 2048 pairs of equivalent equations where the IDEAS domain reasoner could not diagnose the use of a strategy.

*Data analysis and result*: We let our system calculate a diagnosis for each of the 2048 requests. This took on average .17 seconds where the longest computing time was .55 seconds. Of the 2048 student steps, 1749 were diagnosed as strategy applications where a student combined steps, and the remaining 299 steps received a strategy deviation diagnosis. To evaluate these diagnoses, we took random samples of 70 unique steps diagnosed as strategy deviations, and of 70 unique steps diagnosed as strategy applications. The first and third author, both experienced mathematics teachers, assessed the steps where they coded them as either strategy application or strategy deviation. Their interrater reliability was $\kappa = .97$ and after a discussion, they fully agreed. Comparing the teacher coding and the system diagnosis revealed alignment for all 140 strategy steps.

## 5    Discussion of the PT design

Our PT design consists of three principles: (a) defining an expert strategy, (b) identifying variations on expressions in a model solution, and (c) describing variations in terms of relations. The PT design has similarities with both MT and CBM, the expert strategy is defined using MT and the variations are then defined through constraints. Could the same results be achieved by using either one of these approaches? On the one hand using only MT for multistep diagnoses would lead to combinatorial explosion making diagnosis infeasible with limited resources. While on the other hand using only constraints would reduce cognitive fidelity making it hard to implement such a multistep diagnosis system. The PT approach provides a balance between cognitive fidelity and feasibility to efficiently diagnose student strategies. Moreover, PT can provide a strategy hint when a student follows the desired strategy and a diagnosis based on a violated relation when a student deviates from the desired strategy.

A PT diagnosis of a student strategy can be used to provide feedback about the current input. For instance, for a correct strategy step a green checkmark can be displayed. For equivalent steps that deviate from the intended strategy, a yellow checkmark can be displayed together with a message based on the constraint that was violated. A diagnosis provides information on the extent to which a student employs a certain strategy even if the student combines several steps. When a certain strategy is detected within a task solution it can be added to the student model. Hence, PT can contribute to constructing a student model.

How generalizable is the PT approach? For each of the design principles above we argue that it can be generalized to various domains. (a) *Defining an expert strategy*: Heeren & Jeuring (2017) describe how general problem-solving procedures can be modeled through their strategy DSL. The DSL is general and independent of the task domain and has been used for many problem-solving procedures. (b) *Identifying variations*: Heeren & Jeuring (2009) describe normal forms and argue that their techniques are applicable to various knowledge domains. Gerdes et al. (2012) use normal forms for comparing variations of student input in a programming tutor. Furthermore, Mitrovic (2012) states that constraints can capture what is known about a domain. As such, mandatory parts of a student input can be evaluated, while allowing for multiple representations differing only in non-essential elements. (c) *Describing variations in terms of relations*: the use of relations can be considered an extension to CBM. In the ordering of the relations, previous relations serve as relevance conditions for the current relation which can be seen as a satisfaction condition. Mitrovic (2012) shows that CBM can be, and has been, used to model knowledge over a wide variety of domains.

Combining the three principles above to model procedural knowledge, we see that generally a DSL-expression can model an *ideal* problem-solving procedure, giving a model solution. Possible variations on an ideal procedure differ in non-essential elements and are modeled by constraints expressed as relations and normal forms. Ordering constraints such that each constraint is relevant for the subsequent constraints, makes multi-step diagnoses of the procedure manageable.

## 6    Conclusion and future work

Here we answer our research question on the design of a service to provide a strategy diagnosis when a student combines several steps. From our implementation for polynomial equations, we conclude that PT is an adequate answer in this domain. Through PT, all strategy diagnoses agreed with teacher diagnoses and were computed in at most .55 seconds per diagnosis.

We provided arguments for the generalizability of the approach. Although we showed the validity of strategy deviations PT diagnoses, the validity of a specific diagnosis of a violated relation is not studied in this paper and deserve further attention. In the near future we plan to study whether students can benefit from PT strategy feedback through classroom experiments.

**Acknowledgments** We thank Paul Drijvers for his contributions to the study as a whole. Furthermore, we especially thank Bastiaan Heeren, who instead of showing us how to solve problems, taught us how to solve problems.

## References

1. Anderson, J. R., Boyle, C. F., Corbett, A. T., & Lewis, M. W. (1990). *Cognitive modeling and intelligent tutoring*. https://doi.org/10.7551/mitpress/1167.003.0002

2. Anderson, J. R., Corbett, A. T., Koedinger, K. R., & Pelletier, R. (1995). Cognitive Tutors: Lessons Learned. *Journal of the Learning Sciences*, *4*(2), 167–207. https://doi.org/10.1207/s15327809jls0402_2
3. Baral, S., Worden, E., Lim, W.-C., Luo, Z., Santorelli, C., Gurung, A., & Heffernan, N. (2024). Automated Feedback in Math Education: A Comparative Analysis of LLMs for Open-Ended Responses. *ArXiv Preprint ArXiv:2411.08910*.
4. Bokhove, C. (2017). Using technology for digital mathematics textbooks: More than the sum of the parts. *International Journal for Technology in Mathematics Education*, *24*(3), 107–114. https://doi.org/10.1564/tme_v24.3.01
5. Bokhove, C., & Drijvers, P. (2012). Effects of feedback in an online algebra intervention. *Technology, Knowledge and Learning*, *17*, 43–59. https://doi.org/10.1007/s10758-012-9191-8
6. Gerdes, A., Heeren, B., Jeuring, J., & van Binsbergen, L. T. (2017). Ask-Elle: an Adaptable Programming Tutor for Haskell Giving Automated Feedback. *International Journal of Artificial Intelligence in Education*, *27*(1), 65–100. https://doi.org/10.1007/s40593-015-0080-x
7. Gerdes, A., Jeuring, J., & Heeren, B. (2012). An interactive functional programming tutor. *Annual Conference on Innovation and Technology in Computer Science Education, ITiCSE*. https://doi.org/10.1145/2325296.2325356
8. Heeren, B., & Jeuring, J. (2009). Canonical forms in interactive exercise assistants. *International Conference on Intelligent Computer Mathematics*, 325–340. https://doi.org/10.1007/978-3-642-02614-0_27
9. Heeren, B., & Jeuring, J. (2014). Feedback services for stepwise exercises. *Science of Computer Programming*, *88*, 110-129. https://doi.org/10.1016/j.scico.2014.02.021
10. Heeren, B., & Jeuring, J. (2017). An extensible domain-specific language for describing problem-solving procedures. Artificial Intelligence in Education: 18th International Conference, AIED 2017, Wuhan, China, June 28–July 1, 2017, Proceedings 18, 77–89. https://doi.org/10.1007/978-3-319-61425-0_7
11. Heeren, B., & Jeuring, J. (2020). *Automated feedback for mathematical learning environments*. Universität Duisburg-Essen.
12. Heeren, B., Jeuring, J., & Gerdes, A. (2010). Specifying rewrite strategies for interactive exercises. *Mathematics in Computer Science*, *3*, 349–370. https://doi.org/10.1007/978-3-642-02614-0_27
13. Jia, J., Wang, T., Zhang, Y., & Wang, G. (2024). The comparison of general tips for mathematical problem solving generated by generative AI with those generated by human teachers. *Asia Pacific Journal of Education*, *44*(1), 8–28. https://doi.org/10.1080/02188791.2023.2286920
14. Mitrovic, A. (2012). Fifteen years of constraint-based tutors: what we have achieved and where we are going. *User Modeling and User-Adapted Interaction*, *22*, 39–72. https://doi.org/10.1007/s11257-011-9105-9
15. Mitrovic, A., Koedinger, K. R., & Martin, B. (2003). A comparative analysis of cognitive tutoring and constraint-based modeling. *International Conference on User Modeling*, 313–322. https://doi.org/10.1007/3-540-44963-9_42
16. Ohlsson, S., & Mitrovic, A. (2007). Fidelity and Efficiency of Knowledge Representations for Intelligent Tutoring Systems. *Technology, Instruction, Cognition & Learning*, *5*(2).
17. Roll, I., Aleven, V., & Koedinger, K. R. (2010). The invention lab: Using a hybrid of model tracing and constraint-based modeling to offer intelligent support in inquiry environments. *Intelligent Tutoring Systems: 10th International Conference, ITS 2010, Pittsburgh, PA,*

*USA, June 14-18, 2010, Proceedings, Part I 10*, 115–124. https://doi.org/10.1007/978-3-642-13388-6_16
18. Sangwin, C. (2013). *Computer aided assessment of mathematics*. OUP Oxford. https://doi.org/10.1093/acprof:oso/9780199660353.001.0001
19. Sangwin, C. (2015). Computer Aided Assessment of Mathematics Using STACK. In *Selected Regular Lectures from the 12th International Congress on Mathematical Education* (pp. 695–713). https://doi.org/10.1007/978-3-319-17187-6_39
20. Tacoma, S., Heeren, B., Jeuring, J., & Drijvers, P. (2020). Intelligent feedback on hypothesis testing. *International Journal of Artificial Intelligence in Education*, *30*(4), 616–636. https://doi.org/10.1007/s40593-020-00218-y
21. Van der Hoek, G., Heeren, B., Bos, R., Drijvers, P., & Jeuring, J. (2024). Students' experiences with automated final answer diagnoses for mathematics tasks. In P. Iaonnone, F. Moons, C. Drüke-Noe, E. Geraniou, F. Morselli, K. Klingbeil, M. Veldhuis, S. Olsher, H. Corinna, & Peter. Gonscherowski (Eds.), *FAME 1 – Feedback & Assessment in Mathematics Education (ETC 14)* (pp. 293–301). Utrecht University and ERME.
22. Willsey, M., Nandi, C., Wang, Y. R., Flatt, O., Tatlock, Z., & Panchekha, P. (2021). Egg: Fast and extensible equality saturation. *Proceedings of the ACM on Programming Languages*, *5*(POPL), 1–29. https://doi.org/10.1145/3434304